\documentclass[letterpaper]{article} 
\PassOptionsToPackage{table}{xcolor}
\usepackage{xcolor}
\usepackage{colortbl}
\usepackage{aaai2026}  
\usepackage{times}  
\usepackage{helvet}  
\usepackage{courier}  
\usepackage[hyphens]{url}  
\usepackage{graphicx} 
\usepackage{natbib}  
\usepackage{caption} 
\usepackage{algorithm}
\usepackage{algorithmic}
\usepackage{amsmath}
\usepackage{amssymb}
\usepackage{pifont}
\usepackage[table]{xcolor}
\usepackage{booktabs}
\usepackage{multirow}
\usepackage{makecell}

\usepackage{enumitem}

\usepackage{newfloat}
\usepackage{listings}
\DeclareCaptionStyle{ruled}{labelfont=normalfont,labelsep=colon,strut=off} 
\floatstyle{ruled}
\newfloat{listing}{tb}{lst}{}
\floatname{listing}{Listing}
\title{Depth3DLane: Monocular 3D Lane Detection via Depth Prior Distillation}

\author{
    Dongxin Lyu\textsuperscript{\rm 1*},
    Han Huang\textsuperscript{\rm 2*},
    Cheng Tan\textsuperscript{\rm 3},
    Zimu Li\textsuperscript{\rm 4}
}

\affiliations {
    \textsuperscript{\rm 1}Jilin University, lvdx2122@mails.jlu.edu.cn\\
    \textsuperscript{\rm 2}Fuzhou University, China, 832201324@fzu.edu.cn\\
    \textsuperscript{\rm 3}Zhejiang University \& Westlake University, tancheng@westlake.edu.cn\\
    \textsuperscript{\rm 4}Shanghai Jiao Tong University, li-zimu@sjtu.edu.cn
}

\begin{document}

\maketitle

\renewcommand{\thefootnote}{\fnsymbol{footnote}}
\footnotetext[1]{Equal Contribution.}

\begin{abstract}
Monocular 3D lane detection is challenging due to the difficulty in capturing depth information from single-camera images. A common strategy involves transforming front-view (FV) images into bird's-eye-view (BEV) space through inverse perspective mapping (IPM), facilitating lane detection using BEV features. However, IPM's flat-ground assumption and loss of contextual information lead to inaccuracies in reconstructing 3D information, especially height. To address these limitations in a principled manner, we introduce Depth3DLane, a framework that systematically resolves ambiguity through a three-stage pipeline. First, our framework deconstructs the ambiguous BEV representation using a Depth Structure Deconstruction Head, which explicitly models height variations to alleviate the flat-ground assumption. Next, a Contextual Prior Injection module resolves ambiguities by distilling and injecting high-level 3D structural priors from a powerful teacher model. Finally, a Geometric Coherence Refiner refines the output by enforcing the smoothness and continuity inherent in physical roads. Extensive experiments show that our method achieves an F1-score of 98.9\% on ApolloSim with a real-time speed of 88 FPS, and remains highly competitive on OpenLane, surpassing previous state-of-the-art methods in overall performance.

\end{abstract}

\begin{links}
    \link{Code}{https://anonymous.4open.science/r/Depth3DLane-DCDD}
\end{links}


\section{Introduction}
\label{sec:intro}

Monocular 3D lane detection represents a fundamental challenge in autonomous driving, where accurate spatial understanding from single-camera inputs is crucial for vehicle localization, navigation, and safety-critical decision making. While cost-effective compared to multi-sensor systems, monocular approaches face the inherent ambiguity of depth estimation from 2D projections. This challenge is particularly pronounced in complex driving scenarios where traditional assumptions fail, such as non-planar road surfaces, varying elevations, and far-range detection where spatial accuracy becomes increasingly critical.

\begin{figure}[h]
  \centering
   \includegraphics[width=1\linewidth]{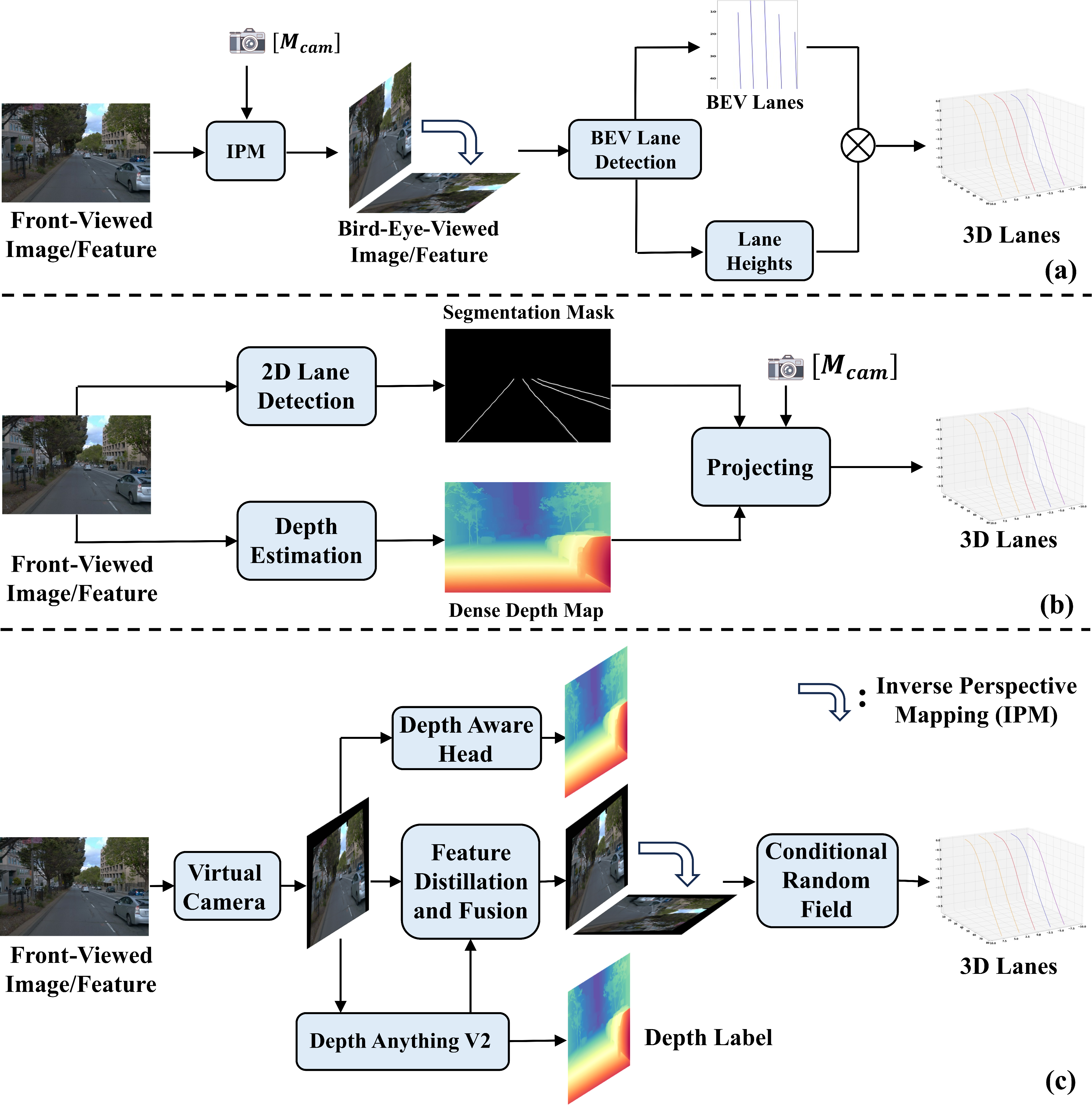}
   \caption{(a) BEV-based methods convert front-view images to BEV for lane detection; (b) Non-BEV methods project 2D lanes into 3D using depth estimation; (c) The proposed Depth3DLane framework systematically resolves ambiguity through three stages: \textit{Depth Structure Deconstruction}, \textit{Contextual Prior Injection}, and \textit{Geometric Coherence Refinement} for coherent 3D lane predictions.}
   \label{fig:onecol}
\end{figure}

A variety of methodologies have emerged to tackle this challenge, which is illustrated in Figure~\ref{fig:onecol}. Early CNN-based approaches such as BEV-LaneDet~\cite{wang2023bev} employ a \textit{Virtual Camera} that transforms front-view images into a consistent BEV representation, adapting effectively to complex scenarios through keypoint-based lane representation. Similarly, Anchor3DLane~\cite{huang2023anchor3dlane} utilizes anchor-based 3D lane detection directly from front-view features, bypassing the need for explicit BEV transformations while maintaining computational efficiency. GroupLane~\cite{li2024grouplane} introduces a classification strategy within the BEV framework, accommodating lanes in multiple orientations and enabling feature interaction across instance groups.

Meanwhile, Transformer-based methods have demonstrated superior accuracy through attention mechanisms and structured representations. PersFormer~\cite{chen2022persformer} extends early 3D lane detection by integrating Transformer-based spatial transformations, enhancing feature robustness across complex perspectives while introducing the foundational OpenLane dataset. Building upon these advances, LATR~\cite{luo2023latr} develops lane-aware query generators and dynamic 3D positional embeddings, leveraging cross-attention mechanisms to iteratively refine lane representations. CurveFormer~\cite{bai2023curveformer} utilizes sparse query representations and cross-attention mechanisms to effectively regress polynomial coefficients of 3D lanes, while CurveFormer++~\cite{bai2024curveformer++} proposes a single-stage detection method that facilitates direct inference from perspective images without requiring view transformations.

Recent developments have focused on addressing specific challenges through specialized innovations. LaneCPP~\cite{Pittner2024LaneCPP} introduces optimized implementations for enhanced real-time performance while maintaining detection accuracy. DV-3DLane~\cite{Luo2024DV3DLane} incorporates depth-aware features to enhance spatial understanding, demonstrating the growing importance of depth information in monocular 3D lane detection. PVALane~\cite{Zheng2024PVALane} combines positional and appearance features through sophisticated attention mechanisms, achieving robust performance across diverse environmental conditions including challenging scenarios such as intersections and extreme weather.

Despite these advances, current methods face three fundamental limitations that compromise their effectiveness. First, the prevalent use of inverse perspective mapping (IPM) introduces systematic errors through its flat-ground assumption, particularly affecting height estimation in non-planar scenarios~\cite{huang2023anchor3dlane, luo2023latr, kim2023d}. Second, existing architectures struggle to capture multi-scale depth relationships essential for accurate 3D spatial reasoning~\cite{garnett20193d, guo2020gen, wang2023bev, liu2022learning, chen2022persformer, bai2023curveformer}. Third, keypoint-based detection methods, while computationally efficient, suffer from spatial discontinuities that violate the inherent geometric constraints of lane structures. Alternative approaches using direct regression~\cite{li2022reconstruct}, weakly supervised learning~\cite{ai2023ws}, or LiDAR integration~\cite{park2024heightlane} either increase system complexity or fail to achieve the precision required for safety-critical applications.

To address these fundamental challenges in a principled manner, we introduce Depth3DLane, a framework that systematically resolves ambiguity through a three-stage pipeline: Deconstruction, Injection, and Refinement. Our main idea is that effective 3D lane detection requires a combined approach that starts with creating a solid geometric foundation, then enriches contextual information, and finally ensures physical realism.
Our contributions are:
\begin{enumerate}
    \item \textbf{Depth Structure Deconstruction (DSD):} The first stage employs a head that actively deconstructs the ambiguous BEV representation by explicitly modeling height variations. This directly alleviates the flat-ground assumption and provides a strong, physically-aware 3D foundation.
    \item \textbf{Contextual Prior Injection (CPI):} The second stage addresses single-image limitations by injecting high-level 3D structural insights. By learning from a powerful teacher model, this module equips our network with the context needed to resolve local ambiguities.
    \item \textbf{Geometric Coherence Refiner (GCR):} The final stage refines the output by ensuring physical plausibility. It captures the natural smoothness and continuity of roads using a \textit{Conditional Random Field (CRF)}, ensuring the final lane structures are consistent and robust.
\end{enumerate}


\section{Related Work}
\label{sec:related_work}
\textbf{2D Lane Detection.} 2D lane detection methods include segmentation-based~\cite{jin2022eigenlanesdatadrivenlanedescriptors, liu2021end}, anchor-based~\cite{li2019line,tabelini2021keep,zheng2022clrnet}, keypoint-based~\cite{wang2022keypoint, qu2021focus, ko2021key}, and curve-based~\cite{tabelini2021polylanenet, van2019end, liu2021end} approaches. While effective in structured environments, these methods require IPM transformations to handle 3D scenarios, limiting their versatility across diverse terrains.

\textbf{Monocular 3D Lane Detection.} Monocular 3D lane detection addresses depth ambiguity in 2D methods by leveraging spatial information from single camera images~\cite{maMonocular3DLane2024}. Early methods like 3D-LaneNet~\cite{garnett20193d} and Gen-LaneNet~\cite{guo2020gen} established anchor-based frameworks but struggled with complex geometric scenarios.

\textbf{CNN-based Methods} emphasize computational efficiency. GroupLane~\cite{li2024grouplane} introduces row-based classification for multi-directional lanes, while recent work includes optimizations in LaneCPP~\cite{Pittner2024LaneCPP} and depth-aware features in DV-3DLane~\cite{Luo2024DV3DLane}.

\textbf{Transformer-based Methods} achieve superior accuracy through attention mechanisms. Beyond foundational work, CurveFormer++~\cite{bai2024curveformer++} enables direct inference from perspective images, while PVALane~\cite{Zheng2024PVALane} demonstrates robust performance across challenging conditions including intersections and extreme weather.
We adopt the \textit{Virtual Camera} concept for viewpoint standardization and enhance keypoint-based detection with our \textit{Geometric Coherence Refiner} to address spatial discontinuity challenges.

\textbf{Monocular Depth Estimation (MDE).} Depth estimation differentiates monocular 3D from 2D lane detection but remains challenging for single-camera systems. Recent advances include Depth Anything~\cite{yang2024depth} and Depth Anything V2~\cite{yang2024depth2}, which enhance generalization through large-scale training and attention mechanisms. Our work leverages Depth Anything V2 through knowledge distillation, transferring depth understanding to enhance 3D lane detection while maintaining real-time efficiency.


\section{Methodology}
\label{sec:methodology}

\begin{figure*}[t]
  \centering
   \includegraphics[width=1\linewidth]{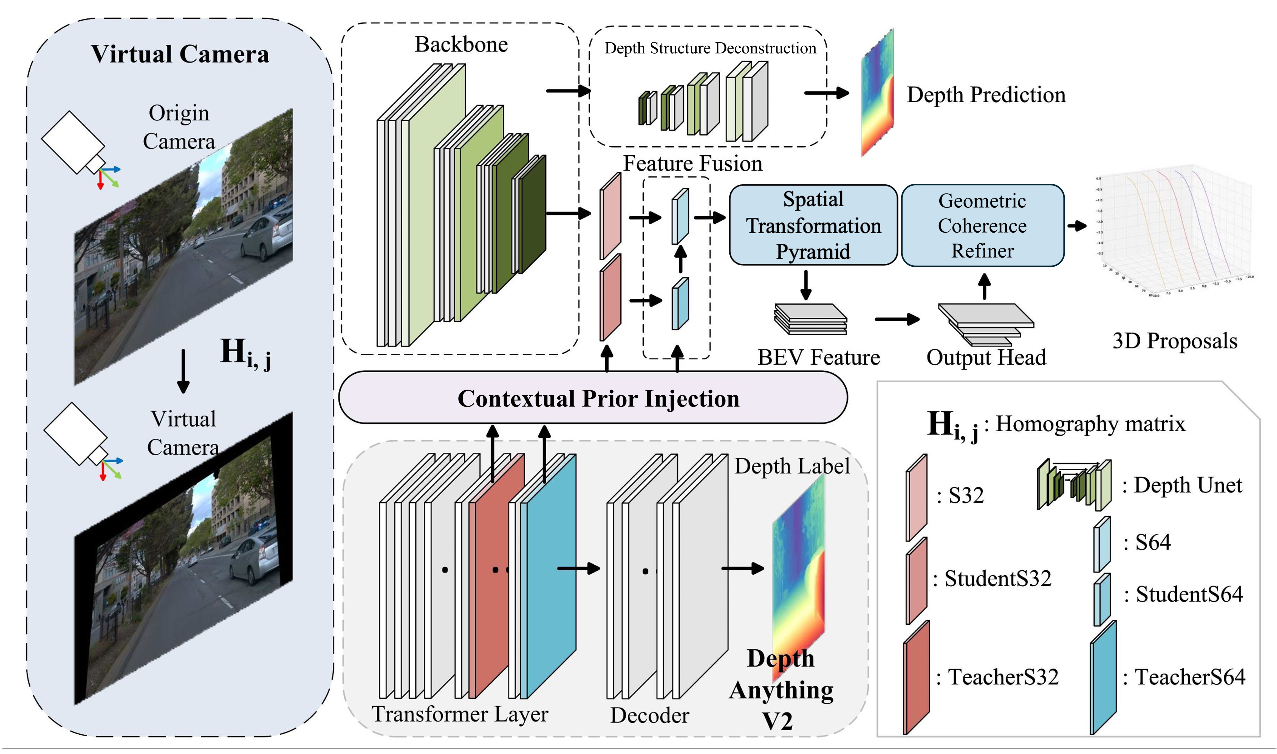}
    \caption{Overview of the proposed Depth3DLane framework, which follows a three-stage pipeline. First, input images are standardized using a \textit{Virtual Camera}~\cite{wang2023bev}. A backbone extracts front-view features. \textbf{Stage 1: Deconstruction}: The proposed Depth Structure Deconstruction (DSD) Head performs multi-scale depth extraction to model an initial 3D height representation. \textbf{Stage 2: Injection}: Depth knowledge is distilled from a pretrained Depth Anything V2~\cite{yang2024depth2} into student modules (StudentS32 and StudentS64). This knowledge is combined with corresponding backbone features, injecting rich contextual information to resolve ambiguities. \textbf{Stage 3: Refinement}: The enriched features are transformed into BEV through a \textit{Spatial Transformation Pyramid}. Finally, our Geometric Coherence Refiner (GCR), implemented as a \textit{Conditional Random Field}, produces robust and consistent 3D lane predictions.}
   \label{fig:structure}
\end{figure*}

The overall structure of our proposed Depth3DLane is shown in Figure \ref{fig:structure}. Given an input image \( \textbf{I} \in \mathbb{R}^{H \times W \times 3} \), its camera parameters are first standardized using a \textit{Virtual Camera}~\cite{wang2023bev}. Our framework then processes the image through a three-stage pipeline: \textit{Depth Structure Deconstruction}, \textit{Contextual Prior Injection}, and \textit{Geometric Coherence Refinement}. This pipeline is trained end-to-end, with each stage guided by a carefully designed loss function to ensure robust and reliable learning. Finally, the predicted 3D lanes are projected onto the road plane \( P_{\text{road}} \) (\(z=0\) in road-ground coordinates), denoted as \( C_{\text{road}} = (x, y, z) \).

\subsection{Stage 1: Depth Structure Deconstruction}

The first stage addresses the problem of retrieving vertical structure from a 2D image, which is often compromised by the flat-ground assumption in IPM-based methods~\cite{huang2023anchor3dlane}. Our Depth Structure Deconstruction (DSD) Head clarifies the ambiguous BEV representation by modeling height variations.

The DSD Head is a U-Net~\cite{ronneberger2015u} inspired encoder-decoder module connected to the feature backbone (\(F_{\text{backbone}}\)). It captures multi-scale 3D spatial details. The encoder path extracts depth-related features at various scales, while an auxiliary decoder path reconstructs a dense depth map during training. This process can be expressed as:
\begin{equation}
D_{\text{initial}} = \text{Decoder}(\text{Encoder}(F_{\text{backbone}}))
\end{equation}
where \( D_{\text{initial}} \) is the reconstructed depth map used for auxiliary supervision. This structure provides a strong supervisory signal, encouraging the network to encode the road's vertical geometry into its features. To optimize this module, we use a combined depth loss \( \mathcal{L}_{\text{depth}} \), which includes the Structural Similarity Index Measure (SSIM)~\cite{wang2004ssim} and Scale-Invariant Logarithmic (SiLog)~\cite{eigen2014depth} losses:
\begin{equation}
\mathcal{L}_{\text{depth}} = \alpha \mathcal{L}_{\text{SSIM}} + \beta \mathcal{L}_{\text{SiLog}}
\end{equation}
Here, \( \mathcal{L}_{\text{SSIM}} \) preserves high-frequency structural details, while \( \mathcal{L}_{\text{SiLog}} \) promotes scale-invariant depth accuracy, with hyperparameters \( \alpha=0.3 \) and \( \beta=1.0 \). This supervision ensures the DSD Head builds a physically aware foundation. The decoder is removed during inference to keep real-time performance.

\subsection{Stage 2: Contextual Prior Injection}

The DSD Head provides an initial 3D estimate. However, a single-image input limits its view. To solve this, our second stage introduces the Contextual Prior Injection (CPI) Module. This module reduces uncertainties by adding high-level 3D structural information from a strong pre-trained teacher model, Depth Anything V2~\cite{yang2024depth2}.

The main function of this module is to fuse these externally-derived "injected features" (\(F_{\text{injected}}\)) with our network's backbone features (\(F_{\text{backbone}}\)) to create a refined, context-aware feature map:
\begin{equation}
F_{\text{fused}} = \text{Concat}(F_{\text{backbone}}, F_{\text{injected}})
\end{equation}

To generate these strong \(F_{\text{injected}}\) features, we strategically distill knowledge from deep layers of the teacher model (layers 17 and 23) into two student modules, StudentS32 and StudentS64. This hierarchical distillation allows the network to capture both fine-grained lane boundaries and broader spatial context. The knowledge transfer is optimized via a feature distillation loss, \( \mathcal{L}_{\text{distill}} \), which aligns the student features with the teacher's representations using both Mean Squared Error (MSE) and Cosine Embedding losses:
\begin{equation}
\mathcal{L}_{\text{distill}} = \lambda_{\text{mse}} \left( \mathcal{L}_{\text{mse}}^{s32} + \mathcal{L}_{\text{mse}}^{s64} \right) + \lambda_{\text{cos}} \left( \mathcal{L}_{\text{cos}}^{s32} + \mathcal{L}_{\text{cos}}^{s64} \right)
\end{equation}
The weights are set to \( \lambda_{\text{mse}}=1.0 \) and \( \lambda_{\text{cos}}=0.5 \) to match both the magnitude and orientation of the teacher's feature representations.

\subsection{Stage 3: Geometric Coherence Refiner}

Keypoint-based detection, while efficient, often results in fragmented or non-smooth lane segments, particularly problematic for safety-critical applications requiring high spatial coherence~\cite{wang2022keypointbasedglobalassociationnetwork}. To address this, our final stage introduces a \textbf{Geometric Coherence Refiner (GCR)}, which enforces the natural smoothness of roads by modeling lane prediction as a structured problem.

In lane detection models, the prediction error for lane lines that are closer to the camera is typically much smaller than for those farther away. To address this discrepancy, we define the baseline pixel of the $i$-th lane as the pixel located at the bottom center of the lane in the predicted image. This baseline pixel serves as a reference point for the $i$-th lane line. To reduce unnecessary computations and mitigate the influence of irrelevant pixels, we propose constructing an independent \textit{Conditional Random Field} for each lane line. Specifically, a separate \textit{Conditional Random Field} is initiated from the baseline pixel of each lane, ensuring a more efficient and localized optimization process.

\begin{figure}
  \centering
   \includegraphics[width=1\linewidth]{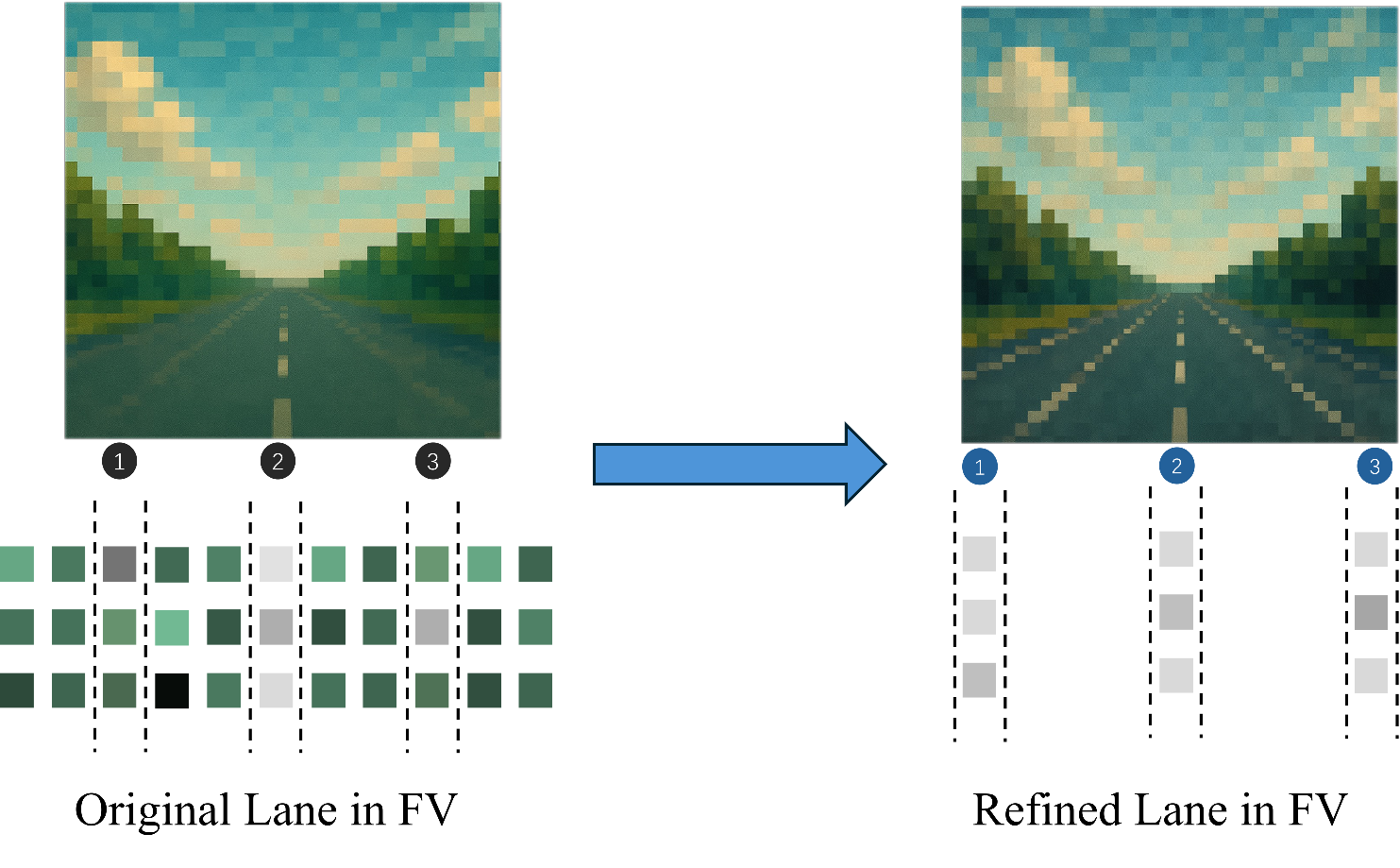}
   \caption{Geometric Coherence Refiner mechanism. An independent CRF is constructed for each lane, initiated from the baseline pixel at the lane bottom. The framework combines unary potentials from model predictions with pairwise potentials incorporating color similarity and depth consistency to enforce spatial coherence and address keypoint fragmentation issues.}
   \label{fig:crf}
\end{figure}

For a prediction matrix from model \textbf{X} where $x_i$ represents the $i$-th pixel
\begin{equation}
E(\textbf{X}) = \sum\limits_{i}\omega_{1}u(x_{i}) + \sum\limits_{i,j}p(x_{i}, x_{j}).
\end{equation}
The unary potential term $u_{x_i}=-logP(x_i)$ considers the predicted probability $P(x_i)$ from the Depth3DLane model.

The pairwise potential in our work consists of two components: color and depth potentials. We use Gaussian kernels to calculate their energies
\begin{equation}
    \begin{aligned}
        p(x_i,x_j)&=\omega_{2}\ p_{color}(x_i,x_j)+\omega_{3}\ p_{depth}(x_i,x_j)\\
        &=\omega_2 exp(-\frac{||\textbf{c}(x_i)-\textbf{c}(x_j)||^2}{2\sigma_{color}^2})\\
        &+\omega_3 exp(-\frac{||d(x_i)-d(x_j)||^2}{2\sigma_{depth}^2}).
    \end{aligned}
\end{equation}
The color potential $p_{color}$ depends on the color distance between adjacent pixels $\textbf{c}(i)-\textbf{c}(j)$ where $\textbf{c}(i)$ denotes the color vector of pixel $i$. The depth potential $p_{depth}$ depends on the output from the Depth Anything V2~\cite{yang2024depth2} model.

\subsection{Prediction and Overall Loss Objective}

The entire network is trained by optimizing a composite loss function that combines a primary lane prediction loss with our auxiliary depth and distillation losses. The primary lane loss, \( \mathcal{L}_{\text{lane}} \), is itself a multi-component objective designed for robust 3D instance prediction, including both 3D and 2D supervisory signals:
\begin{equation}
\begin{aligned}
\mathcal{L}_{\text{lane}}
=&\ \lambda_{\text{conf}}^{3d} \mathcal{L}_{\text{conf}}^{3d}
+ \lambda_{\text{embed}}^{3d} \mathcal{L}_{\text{embed}}^{3d}
+ \lambda_{\text{offset}}^{3d} \mathcal{L}_{\text{offset}}^{3d} \\
&+ \lambda_{\text{height}}^{3d} \mathcal{L}_{\text{height}}^{3d} 
+ \lambda_{\text{seg}}^{2d}   \mathcal{L}_{\text{seg}}^{2d} 
+ \lambda_{\text{embed}}^{2d} \mathcal{L}_{\text{embed}}^{2d}
\end{aligned}
\end{equation}
Here, \( \mathcal{L}_{\text{conf}}^{3d} \) is a confidence loss for lane points, \( \mathcal{L}_{\text{embed}}^{3d} \) is an embedding loss for clustering points into lane instances, and \( \mathcal{L}_{\text{offset}}^{3d} \) and \( \mathcal{L}_{\text{height}}^{3d} \) are regression losses for the BEV offset and height respectively. The 2D losses for segmentation and embedding provide additional supervision. In our experiments, the weights are set to \( \lambda_{\text{conf}}^{3d}=3.0 \), \( \lambda_{\text{embed}}^{3d}=0.5 \), \( \lambda_{\text{offset}}^{3d}=60 \), \( \lambda_{\text{height}}^{3d}=30 \), \( \lambda_{\text{seg}}^{2d}=1.5 \), and \( \lambda_{\text{embed}}^{2d}=0.25 \).

The total loss, \( \mathcal{L}_{\text{total}} \), combines this primary task loss with the auxiliary losses defined previously:
\begin{equation}
\mathcal{L}_{\text{total}}
= \lambda_{\text{lane}}\,\mathcal{L}_{\text{lane}}
+ \lambda_{\text{depth}}\,\mathcal{L}_{\text{depth}}
+ \lambda_{\text{distill}}\,\mathcal{L}_{\text{distill}}
\end{equation}
where the weights \( \lambda_{\text{lane}} \), \( \lambda_{\text{depth}} \), and \( \lambda_{\text{distill}} \) are all set to 1.0 in our experiments to balance the contributions of lane prediction, depth supervision, and feature distillation.


\begin{table*}
    \centering
    \renewcommand{\arraystretch}{1.07} 
    \setlength{\tabcolsep}{1.2mm}

        \begin{tabular}{@{}c|c|ccccc@{}}
        \Xhline{1.5pt}
        \textbf{Scene} & \textbf{Method} & \textbf{F1(\%)}↑ & \textbf{x-Err/N(m)}↓ & \textbf{x-Err/F(m)}↓ & \textbf{z-Err/N(m)}↓ & \textbf{z-Err/F(m)}↓ \\
        \hline
        \hline
        \multirow{7}{*}{\centering Balanced Scene} 
            & PersFormer~\cite{chen2022persformer} & 92.9 & 0.054 & 0.356 & 0.010 & 0.234 \\
            & Anchor3DLane~\cite{huang2023anchor3dlane} & 95.6 & 0.052 & 0.306 & 0.015 & 0.223 \\
            & BEV-LaneDet~\cite{wang2023bev} & 98.7 & \textbf{0.016} & \textbf{0.242} & 0.020 & 0.216 \\
            & LATR~\cite{luo2023latr} & 96.8 & 0.022 & 0.253 & \textbf{0.007} & 0.202 \\
            & LaneCPP~\cite{Pittner2024LaneCPP} & 97.4 & 0.030 & 0.277 & 0.011 & 0.206 \\
            & DV-3DLane~\cite{Luo2024DV3DLane} & 96.4 & 0.046 & 0.299 & 0.016 & 0.213 \\ 
            & \cellcolor{gray!40}Depth3DLane (Ours)
            & \cellcolor{gray!40}\textbf{98.9}
            & \cellcolor{gray!40}0.027
            & \cellcolor{gray!40}0.303
            & \cellcolor{gray!40}0.012
            & \cellcolor{gray!40}\textbf{0.201} \\ 
        \Xhline{1.1pt}
        \multirow{7}{*}{\centering Rarely Observed} 
            & PersFormer~\cite{chen2022persformer} & 87.5 & \textbf{0.107} & 0.782 & 0.024 & 0.602 \\
            & Anchor3DLane~\cite{huang2023anchor3dlane} & 94.4 & 0.094 & 0.693 & 0.027 & 0.579 \\
            & BEV-LaneDet~\cite{wang2023bev} & 99.1 & 0.031 & 0.594 & 0.040 & 0.556 \\
            & LATR~\cite{luo2023latr} & 96.1 & 0.050 & 0.600 & \textbf{0.015} & 0.532 \\
            & LaneCPP~\cite{Pittner2024LaneCPP} & 96.2 & 0.073 & 0.651 & 0.023 & 0.543 \\
            & DV-3DLane~\cite{Luo2024DV3DLane} & 95.6 & 0.071 & 0.664 & 0.025 & 0.568 \\ 
            & \cellcolor{gray!40}Depth3DLane (Ours)
            & \cellcolor{gray!40}\textbf{99.2}
            & \cellcolor{gray!40}0.026
            & \cellcolor{gray!40}\textbf{0.547}
            & \cellcolor{gray!40}0.032
            & \cellcolor{gray!40}\textbf{0.524} \\
        \Xhline{1.1pt}
    \end{tabular}

    \caption{Performance comparison of state-of-the-art methods on the ApolloSim\cite{guo2020gen} dataset across three distinct split settings. "C" and "F" represent close and far ranges, respectively. Our model overall outperforms previous methods, particularly in rare and complex scenarios, demonstrating significant improvements in 3D lane detection accuracy and robustness}
    \label{tab:comparison_on_apollo}
\end{table*}    

\section{Experiments}

\label{sec:experiments}

\subsection{Datasets and Evaluation Metrics}
To evaluate the effectiveness of our work, we conduct experiments on two widely recognized datasets: Apollo Synthetic~\cite{guo2020gen} and OpenLane~\cite{chen2022persformer}.

\textbf{Apollo Synthetic} is a synthetic dataset generated using the Unity 3D engine. It provides photorealistic images across different driving environments such as highways, urban streets, and residential areas. The dataset is particularly useful for controlled testing, with variations in lighting, weather, and road surface conditions, enabling a detailed assessment of lane detection algorithms under a wide array of simulated conditions.

\textbf{OpenLane}, on the other hand, is a large-scale, real-world dataset derived from the Waymo Open Dataset~\cite{sun2020scalability}, consisting of 200,000 frames and over 880,000 annotated lanes. This dataset offers a diverse range of scene metadata, including weather conditions and geographical locations, ensuring a comprehensive evaluation across various environmental factors.

\textbf{Evaluation Metrics}. For evaluation on both 3D datasets, we use the metrics proposed by Gen-LaneNet~\cite{guo2020gen}, which encompass F-Score across different scenes and X/Z errors in various regions. We adopt standard metrics widely used in 3D lane detection research.
\begin{table*}
  \centering
  \renewcommand{\arraystretch}{1.03} 
  \setlength{\tabcolsep}{1.2mm}{



    
     \begin{tabular}{@{}c|c|c|cc|cc|c@{}}
        \Xhline{1.5pt}
        \textbf{Method} & \textbf{F1(\%)}$\uparrow$ & \textbf{Backbone} & \textbf{x-Err/C(m)}$\downarrow$ & \textbf{x-Err/F(m)}$\downarrow$ & \textbf{z-Err/C(m)}$\downarrow$ & \textbf{z-Err/F(m)}$\downarrow$ & \textbf{FPS}$\uparrow $\\
        \hline
        \hline
        PersFormer~\cite{chen2022persformer} & 50.5 & EffNet-B7 & 0.485 & 0.553 & 0.364 & 0.431 & 21\\
        Anchor3DLane~\cite{huang2023anchor3dlane} & 54.3 & EffNet-B3 & 0.275 & 0.310 & 0.105 & 0.135 & 87.3\\
        BEV-LaneDet~\cite{wang2023bev} & 58.4 & ResNet-34 & 0.309 & 0.659 & 0.244 & 0.631 & \textbf{102} \\
        LATR~\cite{luo2023latr} &  61.9 & ResNet-50 & 0.219 & 0.259 & 0.075 & 0.104 & 15.2 \\
        LaneCPP~\cite{Pittner2024LaneCPP} & 60.3 & EffNet-B7 & 0.264 & 0.310 & 0.077 & 0.117 & - \\
        PVALane~\cite{Zheng2024PVALane} & 62.7 & ResNet-50 & 0.232 & 0.259 & 0.092 & 0.118 & 53 \\
        PVALane~\cite{Zheng2024PVALane} & 63.4 & Swin-B & 0.226 & 0.257 & 0.093 & 0.119 & 31 \\
        \Xhline{1.1pt}

        \rowcolor{gray!40} 
        Depth3DLane (Ours) & 61.5 & ResNet-34 & 0.254 & 0.232 & 0.084 & 0.105 & 88\\
        \rowcolor{gray!40} 
        Depth3DLane (Ours) & 62.9 & ResNet-50 & 0.233 & 0.221 & 0.072 & 0.097 & 60\\
        \rowcolor{gray!40} 
        Depth3DLane (Ours) & \textbf{64.7} & Swin-B & \textbf{0.197} & \textbf{0.157} & \textbf{0.069} & \textbf{0.094} & 42\\
        \Xhline{1.5pt}
    \end{tabular}}

\caption{Comparison of different 3D lane detection methods on the OpenLane dataset across various metrics. Our method achieves the highest accuracy and lowest localization errors, indicating superior lane detection performance.}
\label{tab:openlane-general}
\end{table*}
\begin{table*}
  \centering
  \renewcommand{\arraystretch}{1.03} 
  \setlength{\tabcolsep}{1.2mm}{

    \begin{tabular}{@{}cccccccc@{}}
        \Xhline{1.5pt}
        \textbf{Method} & \textbf{Backbone} & \textbf{Up \& Down} & \textbf{Curve} & \textbf{Extreme Weather} & \textbf{Night} & \textbf{Intersection} & \textbf{Merge \& Split} \\
        \hline
        \hline
        PersFormer~\cite{chen2022persformer} & EffNet-B7 & 42.4 & 55.6 & 48.6 & 46.6 & 40.0 & 50.7 \\
        BEV-LaneDet~\cite{wang2023bev} & ResNet-34 & 48.7 & 63.1 & 53.4 & 53.4 & 50.3 & 53.7 \\
        PersFormer~\cite{chen2022persformer} & ResNet-50 & 46.4 & 57.9 & 52.9 & 47.2 & 41.6 & 51.4 \\

        LATR~\cite{luo2023latr} & ResNet-50 & 55.2 & 68.2 & 57.1 & 55.4 & 52.3 & 61.5 \\
        LaneCPP~\cite{Pittner2024LaneCPP} & EffNet-B7 & 53.6 & 64.4 & 56.7 & 54.9 & 52.0 & 58.7 \\
        PVALane~\cite{Zheng2024PVALane} & ResNet-50 & 54.1 & 67.3 & \textbf{62.0} & 57.2 & 53.4 & 60.0 \\

        \Xhline{1.1pt}
        
        \rowcolor{gray!40} 
        Depth3DLane (Ours) & ResNet-34 & 55.7 & 68.8 & 55.7 & 56.2 & 53.1 & 61.4 \\
        \rowcolor{gray!40} 
        Depth3DLane (Ours) & ResNet-50 & 56.8 & 69.5 & 57.2 & 57.4 & 53.8 & 62.9 \\
        \rowcolor{gray!40} 
        Depth3DLane (Ours) & Swin-B & \textbf{58.4} & \textbf{71.3} & 60.7 & \textbf{61.2} & \textbf{55.4} & \textbf{65.0} \\

        \Xhline{1.5pt}
   \end{tabular}}

  \caption{Per-scenario F1 score comparison on the OpenLane dataset, highlighting the robustness of our model across challenging conditions such as intersections, night-time, and extreme weather.}
  \label{tab:openlane-special}
\end{table*}

\subsection{Comparison Studies}
We compare Depth3DLane with recent state-of-the-art methods, including 3D-LaneNet~\cite{garnett20193d}, Gen-LaneNet~\cite{guo2020gen}, PersFormer~\cite{chen2022persformer}, Anchor3DLane~\cite{huang2023anchor3dlane}, and BEV-LaneDet~\cite{wang2023bev}. Evaluation is performed on both OpenLane and Apollo 3D Synthetic datasets under diverse driving conditions, including curved roads, uphill/downhill gradients, and extreme weather scenarios.

\subsubsection{Results on Apollo 3D Synthetic}

As presented in Table~\ref{tab:comparison_on_apollo}, our model attains an F1 score of 98.9\% in the balanced scene and the lowest far-range height error (z-Err/F) of 0.201m among the compared methods, indicating strong performance at long distances. While near-range and far-range lateral accuracy (x-Err/N) are both slightly higher than BEV-LaneDet, the overall F1 and far-range z-axis localization remain competitive and robust in this split.

In the rare subset, our model achieves an F1 score of 99.2\%, together with the lowest far-range errors (x-Err/F = 0.547m, z-Err/F = 0.524m) among the evaluated methods. These results indicate improved robustness under infrequent and complex configurations, particularly at longer viewing distances where depth ambiguity is more pronounced.

\subsubsection{Results on OpenLane}
Table \ref{tab:openlane-general} summarizes comparisons across multiple metrics. Our method achieves leading F1 scores and the lowest x- and z-axis errors across backbones: the Swin-B variant attains the highest F1 of 64.7\%, while the ResNet-50 and ResNet-34 variants reach 62.9\% and 61.5\%, respectively. In addition, the ResNet-34 model delivers strong efficiency (88 FPS) with x-Err/F = 0.232m and z-Err/F = 0.105m, and the Swin-B model achieves x-Err/F = 0.157m and z-Err/F = 0.094m. These results surpass BEV-LaneDet in F1 and far-range localization, reflecting improved robustness in depth and lateral positioning on OpenLane

Additionally, we performed a detailed breakdown of F1 scores across different scenarios within the OpenLane dataset, as shown in Table \ref{tab:openlane-special}. Our method consistently outperformed previous approaches, particularly in challenging environments such as curves, intersections, and night-time conditions. This enhanced performance can be attributed to our method's improved capability to capture spatial relationships and effectively generalize across diverse environmental conditions. By addressing limitations found in prior methods and refining feature extraction through our model, our approach delivers a more robust and precise solution for monocular 3D lane detection in real-world driving scenarios.

\subsection{Ablation Studies}

To validate the effectiveness of our proposed components, we conduct a series of ablation studies on the OpenLane dataset, selecting ResNet-34 \cite{he2016deep} as the backbone for the experiments. These studies evaluate different scale combinations for distillation and assess the individual and combined contributions of the \textit{Depth Structure Deconstruction}, \textit{Contextual Prior Injection}, and \textit{Geometric Coherence Refiner} modules.

\subsubsection{Multi-Scale Feature Alignment for Distillation}

We investigate the impact of aligning features at different scales during the distillation process. Specifically, we experiment with aligning features at multiple scales to enhance the transfer of depth-related information across different levels of abstraction. The results, summarized in Table \ref{tab:scale_combinations}, demonstrate that multi-scale feature alignment significantly boosts model performance.

\begin{figure*}
  \centering
   \includegraphics[width=1\linewidth]{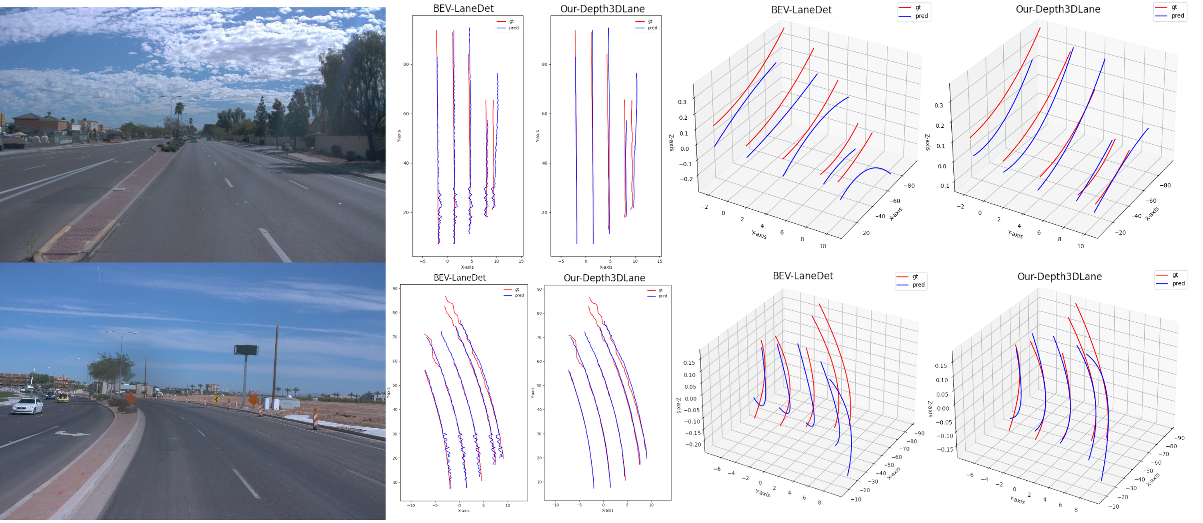}
    \caption{Qualitative comparison between our proposed Depth3DLane and BEV-LaneDet \cite{wang2023bev} on the OpenLane dataset. The first column shows the input images; the second column displays results from BEV-LaneDet in the Bird’s-Eye View (BEV); the third column illustrates our Depth3DLane model’s results in BEV; the fourth column presents BEV-LaneDet’s results in 3D space; and the fifth column depicts our model’s predictions in 3D space. It can be clearly observed that our Depth3DLane model significantly improves lane detection accuracy and continuity by effectively leveraging depth information for enhanced spatial perception, notably reducing localization errors, especially in the z-axis at far distances.}
   \label{fig:banner}
\end{figure*}

\begin{table}
    \centering
    \renewcommand{\arraystretch}{1.07} 
    \setlength{\tabcolsep}{0.8mm}{

    \begin{tabular}{@{}l|ccc@{}}
        \Xhline{1.5pt}
        \textbf{Combination of scales} & \textbf{F-Score} & \textbf{z-Err/C(m)} & \textbf{z-Err/F(m)} \\
        \hline
        \hline
        S16 & 59.0 & 0.106 & 0.137 \\
        S32 & 59.7 & 0.097 & 0.123 \\
        S64 & 59.1 & 0.096 &  0.128\\
        S128 & 58.4 & 0.099 & 0.117 \\
        S32+S64 & \textbf{61.5} & \textbf{0.084} & 0.105 \\
        S32+S64+S128 & 60.9 & 0.091 & \textbf{0.103} \\
        \Xhline{1.5pt}
    \end{tabular}}

    \caption{Ablation on multi-scale feature combinations for depth distillation. S$N$ indicates features downsampled by a factor of $N$. The S32+S64 combination yields the best F-Score.}
    \label{tab:scale_combinations}
\end{table}

The experiments explore the following scale combinations:

\begin{itemize} 
\item \textbf{Single Scale (S16, S32, S64, S128)}: Distilling depth priors from individual feature scales results in varying levels of improvement. For example, using the S32 scale yields the highest performance among single scales with an F-Score of 59.7. 
\item \textbf{Two-Scale Combination (S32 + S64)}: Combining features from the S32 and S64 scales achieves the best F-Score of 61.5 and minimizes z-Error to 0.105, suggesting that deeper layers' features are complementary to mid-level feature representations. 
\item \textbf{Three-Scale Combination (S32 + S64 + S128)}: Adding the S128 scale to the S32 and S64 combination yields an F-Score of 60.9, with a slight increase in z-Error but a significant improvement in angular precision (z-Err/F(m) = 0.103). 
\end{itemize}

The multi-scale fusion approach demonstrates that combining features from different layers at varying depths can effectively capture both fine-grained details and more abstract depth representations, leading to improved overall depth estimation performance.

\begin{table}
    \centering
    \renewcommand{\arraystretch}{1.07} 
    \setlength{\tabcolsep}{1.0mm}{
    

    \begin{tabular}{@{}p{10mm}cc|cccc@{}}
        \Xhline{1.5pt}
        \textbf{DSD} & \textbf{CPI} & \textbf{GCR} & \textbf{F-Score} & \textbf{z-Err/C} & \textbf{z-Err/F} & \textbf{FPS} \\
        \hline
        \hline
        \checkmark &  &  & 60.9 & 0.094 & 0.120 & \textbf{102} \\
         & \checkmark &  & 60.6 & 0.098 & 0.128 & 93 \\
         \checkmark & \checkmark &  & 61.3 & 0.087 & 0.112 & 93 \\
         \checkmark &  & \checkmark & 61.2 & 0.093 & 0.117 & 97 \\
         & \checkmark & \checkmark & 60.9 & 0.091 & 0.115 & 88 \\
         \checkmark & \checkmark & \checkmark & \textbf{61.5} & \textbf{0.084} & \textbf{0.105} & 88 \\
        \Xhline{1.5pt}
    \end{tabular}}
    
    \caption{Ablation study results for different module combinations. This study evaluates the impact of the \textit{Depth Structure Deconstruction (DSD)}, \textit{Contextual Prior Injection (CPI)}, and \textit{Geometric Coherence Refiner (GCR)} modules, evaluated by depth errors in close (z-Err/C) and far (z-Err/F) regions (m) and inference speed (FPS).
    }
    \label{tab:overview}
\end{table}

\subsubsection{Three modules combination}

To further understand the proposed modules, we evaluate the contributions of the \textit{Depth Structure Deconstruction (DSD)}, \textit{Contextual Prior Injection (CPI)}, and \textit{Geometric Coherence Refiner (GCR)} modules within the proposed framework. As illustrated in Table \ref{tab:overview}, we analyze their isolated and combined impacts:
\begin{itemize}
    \item \textbf{DSD}: Enhances depth feature extraction, achieving the highest standalone improvement with an F1-Score of 60.9, significantly improving spatial awareness across varying road geometries.
    \item \textbf{CPI}: Effectively transfers semantic depth knowledge and captures richer contextual information, reaching an F1-Score of 60.6 and showing its importance in depth feature refinement.
    \item \textbf{DSD + CPI}: Achieves robust depth alignment and complementary feature integration, resulting in an increased F1-Score of 61.3. This clearly demonstrates the importance of combining hierarchical depth extraction and semantic feature distillation.
    \item \textbf{DSD + CPI + GCR}: Results in the highest overall performance with an F1-Score of 61.5, further reducing errors and underscoring the effectiveness of spatial coherence enforced by the GCR module. 
\end{itemize}

Overall, these results confirm the critical role and complementarity of the proposed modules, validating the effectiveness of our integration strategy.


\section{Conclusion}
\label{sec:conclusion}

We propose Depth3DLane, a simple yet effective framework for monocular 3D lane detection. By explicitly modeling vertical road structure, our method effectively mitigates inaccuracies from the common flat-ground assumption. Additionally, it injects rich contextual priors to resolve ambiguities and refines predictions with a Conditional Random Field for geometric continuity. Extensive experiments confirm its state-of-the-art performance, especially in challenging far-distance localization. Our ablation studies underscore the crucial synergy of the \textbf{Deconstruct-Inject-Refine} pipeline, as each stage systematically addresses a key challenge unsolved by prior methods. This principled approach charts a new path toward achieving highly coherent and accurate monocular 3D perception, crucial for the future of autonomous driving.


\bibliography{aaai2026}

\begin{thebibliography}{36}
\providecommand{\natexlab}[1]{#1}

\bibitem[{Ai et~al.(2023)Ai, Ding, Zhao, and Zhong}]{ai2023ws}
Ai, J.; Ding, W.; Zhao, J.; and Zhong, J. 2023.
\newblock WS-3D-Lane: Weakly Supervised 3D Lane Detection With 2D Lane Labels.
\newblock In \emph{2023 IEEE International Conference on Robotics and
  Automation (ICRA)}, 5595--5601. IEEE.

\bibitem[{Bai et~al.(2023)Bai, Chen, Fu, Peng, Liang, and
  Cheng}]{bai2023curveformer}
Bai, Y.; Chen, Z.; Fu, Z.; Peng, L.; Liang, P.; and Cheng, E. 2023.
\newblock Curveformer: 3d lane detection by curve propagation with curve
  queries and attention.
\newblock In \emph{2023 IEEE International Conference on Robotics and
  Automation (ICRA)}, 7062--7068. IEEE.

\bibitem[{Bai et~al.(2024)Bai, Chen, Liang, and Cheng}]{bai2024curveformer++}
Bai, Y.; Chen, Z.; Liang, P.; and Cheng, E. 2024.
\newblock CurveFormer++: 3D Lane Detection by Curve Propagation with Temporal
  Curve Queries and Attention.
\newblock \emph{arXiv preprint arXiv:2402.06423}.

\bibitem[{Chen et~al.(2022)Chen, Sima, Li, Zheng, Xu, Geng, Li, He, Shi, Qiao
  et~al.}]{chen2022persformer}
Chen, L.; Sima, C.; Li, Y.; Zheng, Z.; Xu, J.; Geng, X.; Li, H.; He, C.; Shi,
  J.; Qiao, Y.; et~al. 2022.
\newblock Persformer: 3d lane detection via perspective transformer and the
  openlane benchmark.
\newblock In \emph{European Conference on Computer Vision}, 550--567. Springer.

\bibitem[{Eigen, Puhrsch, and Fergus(2014)}]{eigen2014depth}
Eigen, D.; Puhrsch, C.; and Fergus, R. 2014.
\newblock Depth map prediction from a single image using a multi-scale deep
  network.
\newblock In \emph{Advances in neural information processing systems},
  2366--2374.

\bibitem[{Garnett et~al.(2019)Garnett, Cohen, Pe'er, Lahav, and
  Levi}]{garnett20193d}
Garnett, N.; Cohen, R.; Pe'er, T.; Lahav, R.; and Levi, D. 2019.
\newblock 3d-lanenet: end-to-end 3d multiple lane detection.
\newblock In \emph{Proceedings of the IEEE/CVF International Conference on
  Computer Vision}, 2921--2930.

\bibitem[{Guo et~al.(2020)Guo, Chen, Zhao, Zhang, Miao, Wang, and
  Choe}]{guo2020gen}
Guo, Y.; Chen, G.; Zhao, P.; Zhang, W.; Miao, J.; Wang, J.; and Choe, T.~E.
  2020.
\newblock Gen-lanenet: A generalized and scalable approach for 3d lane
  detection.
\newblock In \emph{Computer Vision--ECCV 2020: 16th European Conference,
  Glasgow, UK, August 23--28, 2020, Proceedings, Part XXI 16}, 666--681.
  Springer.

\bibitem[{He et~al.(2016)He, Zhang, Ren, and Sun}]{he2016deep}
He, K.; Zhang, X.; Ren, S.; and Sun, J. 2016.
\newblock Deep residual learning for image recognition.
\newblock In \emph{Proceedings of the IEEE conference on computer vision and
  pattern recognition}, 770--778.

\bibitem[{Huang et~al.(2023)Huang, Shen, Huang, Ding, Dai, Han, Wang, and
  Liu}]{huang2023anchor3dlane}
Huang, S.; Shen, Z.; Huang, Z.; Ding, Z.-h.; Dai, J.; Han, J.; Wang, N.; and
  Liu, S. 2023.
\newblock Anchor3dlane: Learning to regress 3d anchors for monocular 3d lane
  detection.
\newblock In \emph{Proceedings of the IEEE/CVF Conference on Computer Vision
  and Pattern Recognition}, 17451--17460.

\bibitem[{Jin et~al.(2022)Jin, Park, Jeong, Kwon, and
  Kim}]{jin2022eigenlanesdatadrivenlanedescriptors}
Jin, D.; Park, W.; Jeong, S.-G.; Kwon, H.; and Kim, C.-S. 2022.
\newblock Eigenlanes: Data-Driven Lane Descriptors for Structurally Diverse
  Lanes.
\newblock arXiv:2203.15302.

\bibitem[{Kim et~al.(2023)Kim, Byeon, Ji, and Oh}]{kim2023d}
Kim, N.; Byeon, M.; Ji, D.; and Oh, D. 2023.
\newblock D-3DLD: Depth-Aware Voxel Space Mapping for Monocular 3D Lane
  Detection with Uncertainty.
\newblock In \emph{ICASSP 2023-2023 IEEE International Conference on Acoustics,
  Speech and Signal Processing (ICASSP)}, 1--5. IEEE.

\bibitem[{Ko et~al.(2021)Ko, Lee, Azam, Munir, Jeon, and Pedrycz}]{ko2021key}
Ko, Y.; Lee, Y.; Azam, S.; Munir, F.; Jeon, M.; and Pedrycz, W. 2021.
\newblock Key points estimation and point instance segmentation approach for
  lane detection.
\newblock \emph{IEEE Transactions on Intelligent Transportation Systems},
  23(7): 8949--8958.

\bibitem[{Li et~al.(2022)Li, Shi, Wang, and Cheng}]{li2022reconstruct}
Li, C.; Shi, J.; Wang, Y.; and Cheng, G. 2022.
\newblock Reconstruct from top view: A 3d lane detection approach based on
  geometry structure prior.
\newblock In \emph{Proceedings of the IEEE/CVF Conference on Computer Vision
  and Pattern Recognition}, 4370--4379.

\bibitem[{Li et~al.(2019)Li, Li, Hu, and Yang}]{li2019line}
Li, X.; Li, J.; Hu, X.; and Yang, J. 2019.
\newblock Line-cnn: End-to-end traffic line detection with line proposal unit.
\newblock \emph{IEEE Transactions on Intelligent Transportation Systems},
  21(1): 248--258.

\bibitem[{Li et~al.(2024)Li, Han, Ge, Yang, Yu, Wang, Zhang, and
  Zhao}]{li2024grouplane}
Li, Z.; Han, C.; Ge, Z.; Yang, J.; Yu, E.; Wang, H.; Zhang, X.; and Zhao, H.
  2024.
\newblock Grouplane: End-to-end 3d lane detection with channel-wise grouping.
\newblock \emph{IEEE Robotics and Automation Letters}.

\bibitem[{Liu et~al.(2022)Liu, Chen, Liu, Xiong, and Yuan}]{liu2022learning}
Liu, R.; Chen, D.; Liu, T.; Xiong, Z.; and Yuan, Z. 2022.
\newblock Learning to predict 3d lane shape and camera pose from a single image
  via geometry constraints.
\newblock In \emph{Proceedings of the AAAI Conference on Artificial
  Intelligence}, volume~36, 1765--1772.

\bibitem[{Liu et~al.(2021)Liu, Yuan, Liu, and Xiong}]{liu2021end}
Liu, R.; Yuan, Z.; Liu, T.; and Xiong, Z. 2021.
\newblock End-to-end lane shape prediction with transformers.
\newblock In \emph{Proceedings of the IEEE/CVF winter conference on
  applications of computer vision}, 3694--3702.

\bibitem[{Luo et~al.(2023)Luo, Zheng, Yan, Kun, Zheng, Cui, and
  Li}]{luo2023latr}
Luo, Y.; Zheng, C.; Yan, X.; Kun, T.; Zheng, C.; Cui, S.; and Li, Z. 2023.
\newblock Latr: 3d lane detection from monocular images with transformer.
\newblock In \emph{Proceedings of the IEEE/CVF International Conference on
  Computer Vision}, 7941--7952.

\bibitem[{Luo et~al.(2024)}]{Luo2024DV3DLane}
Luo, Y.; et~al. 2024.
\newblock {DV}-3{DL}ane: End-to-end Multi-modal 3D Lane Detection with
  Dual-view Representation.
\newblock In \emph{International Conference on Learning Representations
  (ICLR)}.

\bibitem[{Ma et~al.(2025)Ma, Qi, Zhao, Zheng, Wang, Liu, Liu, and
  Ma}]{maMonocular3DLane2024}
Ma, F.; Qi, W.; Zhao, G.; Zheng, L.; Wang, S.; Liu, Y.; Liu, M.; and Ma, J.
  2025.
\newblock Monocular 3d lane detection for autonomous driving: Recent
  achievements, challenges, and outlooks.
\newblock \emph{IEEE Transactions on Intelligent Transportation Systems}.

\bibitem[{Park, Seo, and Lim(2024)}]{park2024heightlane}
Park, C.; Seo, E.; and Lim, J. 2024.
\newblock HeightLane: BEV Heightmap guided 3D Lane Detection.
\newblock \emph{arXiv preprint arXiv:2408.08270}.

\bibitem[{Pittner et~al.(2024)}]{Pittner2024LaneCPP}
Pittner, M.; et~al. 2024.
\newblock LaneCPP: Continuous 3D Lane Detection Using Physical Priors.
\newblock In \emph{Proceedings of the IEEE/CVF Conference on Computer Vision
  and Pattern Recognition (CVPR)}, 10639--10648.

\bibitem[{Qu et~al.(2021)Qu, Jin, Zhou, Yang, and Zhang}]{qu2021focus}
Qu, Z.; Jin, H.; Zhou, Y.; Yang, Z.; and Zhang, W. 2021.
\newblock Focus on local: Detecting lane marker from bottom up via key point.
\newblock In \emph{Proceedings of the IEEE/CVF conference on computer vision
  and pattern recognition}, 14122--14130.

\bibitem[{Ronneberger, Fischer, and Brox(2015)}]{ronneberger2015u}
Ronneberger, O.; Fischer, P.; and Brox, T. 2015.
\newblock U-net: Convolutional networks for biomedical image segmentation.
\newblock In \emph{Medical image computing and computer-assisted
  intervention--MICCAI 2015: 18th international conference, Munich, Germany,
  October 5-9, 2015, proceedings, part III 18}, 234--241. Springer.

\bibitem[{Sun et~al.(2020)Sun, Kretzschmar, Dotiwalla, Chouard, Patnaik, Tsui,
  Guo, Zhou, Chai, Caine et~al.}]{sun2020scalability}
Sun, P.; Kretzschmar, H.; Dotiwalla, X.; Chouard, A.; Patnaik, V.; Tsui, P.;
  Guo, J.; Zhou, Y.; Chai, Y.; Caine, B.; et~al. 2020.
\newblock Scalability in perception for autonomous driving: Waymo open dataset.
\newblock In \emph{Proceedings of the IEEE/CVF conference on computer vision
  and pattern recognition}, 2446--2454.

\bibitem[{Tabelini et~al.(2021{\natexlab{a}})Tabelini, Berriel, Paixao, Badue,
  De~Souza, and Oliveira-Santos}]{tabelini2021keep}
Tabelini, L.; Berriel, R.; Paixao, T.~M.; Badue, C.; De~Souza, A.~F.; and
  Oliveira-Santos, T. 2021{\natexlab{a}}.
\newblock Keep your eyes on the lane: Real-time attention-guided lane
  detection.
\newblock In \emph{Proceedings of the IEEE/CVF conference on computer vision
  and pattern recognition}, 294--302.

\bibitem[{Tabelini et~al.(2021{\natexlab{b}})Tabelini, Berriel, Paixao, Badue,
  De~Souza, and Oliveira-Santos}]{tabelini2021polylanenet}
Tabelini, L.; Berriel, R.; Paixao, T.~M.; Badue, C.; De~Souza, A.~F.; and
  Oliveira-Santos, T. 2021{\natexlab{b}}.
\newblock Polylanenet: Lane estimation via deep polynomial regression.
\newblock In \emph{2020 25th International Conference on Pattern Recognition
  (ICPR)}, 6150--6156. IEEE.

\bibitem[{Van~Gansbeke et~al.(2019)Van~Gansbeke, De~Brabandere, Neven,
  Proesmans, and Van~Gool}]{van2019end}
Van~Gansbeke, W.; De~Brabandere, B.; Neven, D.; Proesmans, M.; and Van~Gool, L.
  2019.
\newblock End-to-end lane detection through differentiable least-squares
  fitting.
\newblock In \emph{Proceedings of the IEEE/CVF International Conference on
  Computer Vision Workshops}, 0--0.

\bibitem[{Wang et~al.(2022{\natexlab{a}})Wang, Ma, Huang, Hui, Wang, Qian, and
  Zhang}]{wang2022keypoint}
Wang, J.; Ma, Y.; Huang, S.; Hui, T.; Wang, F.; Qian, C.; and Zhang, T.
  2022{\natexlab{a}}.
\newblock A keypoint-based global association network for lane detection.
\newblock In \emph{Proceedings of the IEEE/CVF Conference on Computer Vision
  and Pattern Recognition}, 1392--1401.

\bibitem[{Wang et~al.(2022{\natexlab{b}})Wang, Ma, Huang, Hui, Wang, Qian, and
  Zhang}]{wang2022keypointbasedglobalassociationnetwork}
Wang, J.; Ma, Y.; Huang, S.; Hui, T.; Wang, F.; Qian, C.; and Zhang, T.
  2022{\natexlab{b}}.
\newblock A Keypoint-based Global Association Network for Lane Detection.
\newblock arXiv:2204.07335.

\bibitem[{Wang et~al.(2023)Wang, Qin, Li, Li, Cao, and Xu}]{wang2023bev}
Wang, R.; Qin, J.; Li, K.; Li, Y.; Cao, D.; and Xu, J. 2023.
\newblock Bev-lanedet: An efficient 3d lane detection based on virtual camera
  via key-points.
\newblock In \emph{Proceedings of the IEEE/CVF Conference on Computer Vision
  and Pattern Recognition}, 1002--1011.

\bibitem[{Wang et~al.(2004)Wang, Bovik, Sheikh, and Simoncelli}]{wang2004ssim}
Wang, Z.; Bovik, A.~C.; Sheikh, H.~R.; and Simoncelli, E.~P. 2004.
\newblock Image quality assessment: from error visibility to structural
  similarity.
\newblock \emph{IEEE transactions on image processing}, 13(4): 600--612.

\bibitem[{Yang et~al.(2024{\natexlab{a}})Yang, Kang, Huang, Xu, Feng, and
  Zhao}]{yang2024depth}
Yang, L.; Kang, B.; Huang, Z.; Xu, X.; Feng, J.; and Zhao, H.
  2024{\natexlab{a}}.
\newblock Depth anything: Unleashing the power of large-scale unlabeled data.
\newblock In \emph{Proceedings of the IEEE/CVF Conference on Computer Vision
  and Pattern Recognition}, 10371--10381.

\bibitem[{Yang et~al.(2024{\natexlab{b}})Yang, Kang, Huang, Zhao, Xu, Feng, and
  Zhao}]{yang2024depth2}
Yang, L.; Kang, B.; Huang, Z.; Zhao, Z.; Xu, X.; Feng, J.; and Zhao, H.
  2024{\natexlab{b}}.
\newblock Depth Anything V2.
\newblock \emph{arXiv preprint arXiv:2406.09414}.

\bibitem[{Zheng et~al.(2022)Zheng, Huang, Liu, Tang, Yang, Cai, and
  He}]{zheng2022clrnet}
Zheng, T.; Huang, Y.; Liu, Y.; Tang, W.; Yang, Z.; Cai, D.; and He, X. 2022.
\newblock Clrnet: Cross layer refinement network for lane detection.
\newblock In \emph{Proceedings of the IEEE/CVF conference on computer vision
  and pattern recognition}, 898--907.

\bibitem[{Zheng et~al.(2024)}]{Zheng2024PVALane}
Zheng, Z.; et~al. 2024.
\newblock PVALane: Prior-Guided 3D Lane Detection with View-Agnostic Feature
  Alignment.
\newblock In \emph{Proceedings of the AAAI Conference on Artificial
  Intelligence}, volume~38, 7597--7604. AAAI Press.

\end{thebibliography}


\end{document}